\documentclass[conference]{IEEEtran}
\IEEEoverridecommandlockouts
\usepackage{cite}
\usepackage{amsmath,amssymb,amsfonts}
\usepackage{algorithmic}
\usepackage{graphicx}
\usepackage{textcomp}
\usepackage{xcolor}
\def\BibTeX{{\rm B\kern-.05em{\sc i\kern-.025em b}\kern-.08em
    T\kern-.1667em\lower.7ex\hbox{E}\kern-.125emX}}
\begin{document}

\title{Hybrid Voting-Based Task Assignment in Modular Construction Scenarios\\}

\author{\IEEEauthorblockN{Daniel Weiner}
\IEEEauthorblockA{\textit{Computer Science Department} \\
\textit{Graduate Center, City University of New York}\\
New York, United States \\
dweiner@gradcenter.cuny.edu}
\and
\IEEEauthorblockN{Raj Korpan}
\IEEEauthorblockA{\textit{Computer Science Department} \\
\textit{Hunter College and Graduate Center, City University of New York}\\
New York, United States \\
raj.korpan@hunter.cuny.edu}
}

\maketitle

\begin{abstract}
Modular construction, involving off-site prefabrication and on-site assembly, offers significant advantages but presents complex coordination challenges for robotic automation. Effective task allocation is critical for leveraging multi-agent systems (MAS) in these structured environments. This paper introduces the Hybrid Voting-Based Task Assignment (HVBTA) framework, a novel approach to optimizing collaboration between heterogeneous multi-agent construction teams. Inspired by human reasoning in task delegation, HVBTA uniquely integrates multiple voting mechanisms with the capabilities of a Large Language Model (LLM) for nuanced suitability assessment between agent capabilities and task requirements. The framework operates by assigning Capability Profiles to agents and detailed requirement lists called Task Descriptions to construction tasks, subsequently generating a quantitative Suitability Matrix. Six distinct voting methods, augmented by a pre-trained LLM, analyze this matrix to robustly identify the optimal agent for each task. Conflict-Based Search (CBS) is integrated for decentralized, collision-free path planning, ensuring efficient and safe spatio-temporal coordination of the robotic team during assembly operations. HVBTA enables efficient, conflict-free assignment and coordination, facilitating potentially faster and more accurate modular assembly. Current work is evaluating HVBTA's performance across various simulated construction scenarios involving diverse robotic platforms and task complexities. While designed as a generalizable framework for any domain with clearly definable tasks and capabilities, HVBTA will be particularly effective for addressing the demanding coordination requirements of multi-agent collaborative robotics in modular construction due to the predetermined construction planning involved.

\end{abstract}


\section{Introduction}

The construction industry is increasingly exploring the potential of robotic automation to improve efficiency, safety, and scalability \cite{DAVILADELGADO2019100868}. Modular construction, characterized by the off-site fabrication of components and their subsequent on-site assembly, is particularly well-suited to automated implementation \cite{LIM2022104455}. Modular construction provides a scalable, robot-friendly method for building. Multi-agent systems (MAS) offer significant opportunities for construction tasks ranging from site preparation, such as push manipulation of loose materials for mapping, leveling, and shaping, to the intricate processes of prefabricated assembly \cite{xiao2022recent}.

\begin{figure}
    \centering
    \includegraphics[width=0.5\linewidth]{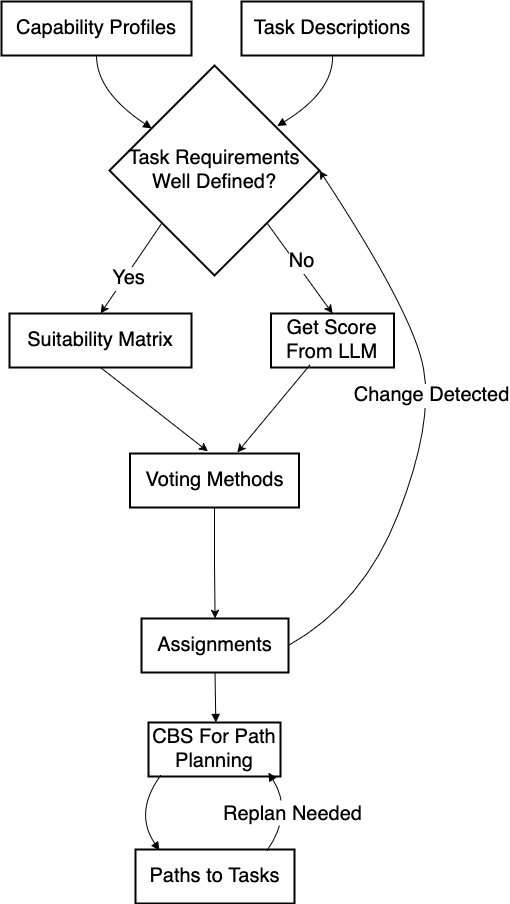}
    \caption{Diagram of the HVBTA system. The Suitability Matrix is created from the Task Descriptions and Capability Profiles when all task requirements are well defined, otherwise, LLM integration is utilized to score suitability, then all scores are passed to voting methods for task allocation. Finally CBS plans paths to the tasks for each agent.}
    \label{fig:1}
\end{figure}

However, effectively deploying and coordinating these MAS, especially when they are heterogeneous (composed of agents with different capabilities), presents substantial challenges. Coordinating a team of heterogeneous agents efficiently requires sophisticated planning and scheduling algorithms. Key tasks include organizing and ordering agent actions and resources, and ensuring seamless agent coordination. Commonly used planning and scheduling techniques include sequencing, bioinspired methods, and optimization. Task assignment, determining which agent should perform which task, is a critical precursor to successful coordination and execution \cite{article}. In dynamic construction environments, agents must not only be assigned tasks appropriately based on their capabilities but also navigate the shared workspace efficiently and without conflict. Traditional task allocation methods may struggle to capture the nuances of agent capabilities and task requirements, deal with unexpected task requirements, or adapt to changing conditions on a construction site \cite{ali2022considerationstaskallocationhumanrobot}.

This paper introduces the Hybrid Voting-Based Task Assignment (HVBTA) framework, as shown in Figure \ref{fig:1}, as a solution to the complex task allocation and coordination challenges in multi-agent modular construction. HVBTA is a generalizable framework that draws inspiration from human reasoning in task delegation. It combines voting with Large Language Model (LLM)-based reasoning to determine which robots are best suited for which tasks. HBVTA is designed to handle well-defined multi-agent settings, making it particularly effective for the demanding coordination requirements of collaborative modular construction due to the predetermined construction planning involved. The next section describes related work, and subsequent sections describe the HVBTA framework in detail and consider how HVBTA may be applied to MAS in the construction sector.

\section{Related Work}
Coordinating heterogeneous multi‑agent teams for structured tasks, such as modular construction, has inspired a range of allocation and planning techniques. Prior work can be grouped into four main paradigms: optimization‑based, market/auction methods, LLM augmentation, and decentralized path planning. In the following, we briefly survey each and highlight how HVBTA builds on and differs from these foundations.

\subsection{Optimization-Based Approaches}
Optimization-based approaches model task allocation as a constrained optimization problem, often using mixed-integer programming (MIP), constraint satisfaction, or heuristic-based models to maximize efficiency metrics such as task coverage, makespan, or resource utilization \cite{CHAKRAA2023104492}. \textbf{Exact methods} (e.g., greedy algorithms, local search, Simulated Annealing) trade solution quality for speed but provide no optimality bounds \cite{CHAKRAA2023104492}. \textbf{Metaheuristics} (e.g., Genetic Algorithms, Particle Swarm Optimization, Ant Colony Optimization) explore large search spaces via population‑based or bio‑inspired strategies, yielding good solutions on mid‑sized problems yet often requiring significant parameter tuning \cite{CHAKRAA2023104492}. \textbf{Hybrid schemes} combine clustering or graph partitioning with exact or metaheuristic solvers to decompose large instances into manageable subproblems \cite{CHAKRAA2023104492}.

While these approaches excel when full cost functions and agent–task compatibilities are known a priori, they often falter under partial, noisy, or semantically rich requirements. In contrast, HVBTA’s voting framework naturally accommodates incomplete suitability data and uses LLM tie‑breaking to handle unseen or context‑dependent demands.

\subsection{Market and Auction‑Based Methods}
Market-based and auction algorithms have long dominated multi-agent task allocation, balancing agent bids against task priorities \cite{taskAllocation1, taskAllocation2}. While these methods are scalable, they often cannot capture the complex, nuanced relationships between tasks and agents that occur in real situations, and may oscillate under dynamic environments, making them less suited to construction environments.

\subsection{LLMs for Semantic Matching} 
Our recent work proposed integrating LLMs to interpret high‑level or unstructured task descriptions and mapping them onto agent capabilities when numeric scores are insufficient in the context of role-playing games \cite{weiner2025hybridvotingbasedtaskassignment}. Some works claim LLMs act as in-context semantic "reasoners" \cite{tang2023largelanguagemodelsincontext}. HVBTA advances this by embedding LLM reasoning directly into its multi‑rule voting pipeline, automatically generating concise prompts to adjudicate between top candidates and feeding the resulting “preference score” back into the assignment.

\subsection{Multi-Agent Path Finding}
Multi‑Agent Pathfinding (MAPF) focuses on finding collision‑free paths for multiple agents moving simultaneously in a shared environment. Classical MAPF methods fall into three broad categories. \textbf{Centralized search} formulates the joint configuration space of all agents and applies global planners (e.g., A*, IDA*) to find an optimal solution. While providing completeness and optimality guarantees, centralized search scales exponentially with the number of agents and is impractical beyond small teams. \textbf{Decoupled or prioritized planning} (e.g., Priority‑Based Search) assigns each agent a priority ordering and plans paths sequentially, treating higher‑priority agents as moving obstacles. These methods run in polynomial time but may fail to find a solution even when one exists, due to priority conflicts. \textbf{Conflict-Based Search} and its bounded‑suboptimal variants (e.g., ECBS) combine the strengths of centralized and decoupled approaches. CBS performs low‑level single‑agent planning independently, then detects pairwise collisions and incrementally adds binary constraints to a high‑level search tree, resolving conflicts until all paths are conflict‑free. This yields optimal or bounded‑suboptimal solutions with dramatically better scalability than naïve centralized search. HVBTA relies on MAPF, in particular CBS, to translate its high‑quality task assignments into collision‑free execution plans. By front‑loading capability‑aware, voting‑driven allocation, HVBTA reduces the number of conflicts that MAPF must resolve, speeding overall coordination.

\section{The HVBTA Framework}

The HVBTA framework represents a novel hybrid approach that integrates structured \textit{Task Descriptions} and \textit{Capability Profiles} with semantic reasoning from LLMs and dynamic path planning from Conflict-Based Search (CBS). While applicable to various domains, its structure is well-suited for the task allocation and coordination needs of multi-agent construction teams. Utilizing well-defined \textit{Task Descriptions} and agent \textit{Capability Profiles}, HVBTA can calculate the suitability of agents to tasks by generating a \textit{Suitability Matrix}, it then votes on which agent should complete which task, and then guides that agent to the task using CBS. HVBTA also accounts for unseen tasks by taking advantage of the reasoning capabilities of a pretrained LLM to provide a suitability score through a previously assembled template prompt, thereby increasing adaptability. The core components of the HVBTA framework are as follows.

\subsection{Task Descriptions and Agent Capability Profiles}
The framework begins by defining tasks and agents. Each construction effort, whether mapping a site, leveling ground using push manipulation, or placing a prefabricated module, is defined by a \textit{Task Description}, a detailed set of requirements outlining what is necessary for successful execution. Concurrently, each agent is described through its \textit{Capability Profile}, detailing its specific abilities, tools (e.g., general-purpose pushing tools), strengths (e.g., precision, lifting capacity, speed), and limitations. This dual representation allows for a rigorous comparison of agent competencies against task demands. Table \ref{table:1} shows an example of two \textit{Task Descriptions} and three \textit{Capability Profiles}.



\begin{table}[t]
\centering
\caption{Simple example of \textit{Task Descriptions} and agent \textit{Capability Profiles}}
\label{table:1}
\begin{tabular}{|l|p{40pt}|p{40pt}|p{40pt}|}
\hline
\textbf{Task}    & \textbf{Payload Size}     & \textbf{Terrain Type}       & \textbf{Reach Needed} \\ \hline
Place Wall Panel & 400 kg                    & Flat                        & 2.0 m                 \\ \hline
Transport Module & 300 kg                    & Uneven                      & 2.5 m                 \\ \hline
                 &                           &                             &                       \\ \hline
\textbf{Agent}   & \textbf{Payload Capacity} & \textbf{Terrain Capability} & \textbf{Reach}        \\ \hline
A                & 500 kg                    & Flat                        & 2.0 m                 \\ \hline
B                & 100 kg                    & Fixed                       & 1.2 m                 \\ \hline
C                & 450 kg                    & Uneven                      & 2.8 m                 \\ \hline
\end{tabular}
\end{table}

\subsection{Suitability Matrix Generation}
HVBTA constructs a quantitative \textit{Suitability Matrix} to evaluate the compatibility between agents and tasks. For every possible agent-task pair, the matrix assigns a score that reflects how well the agent’s capabilities align with the task’s requirements. Scores are calculated using a rule-based approach that compares the agent's \textit{Capability Profile} with the \textit{Task Description}, resulting in higher scores for agents better suited to a task. For example, in Table \ref{table:1}, while the suitability for both \textit{Agent A} and \textit{Agent C} would be very high for \textit{Place Wall Panel}, only \textit{Agent C} has the capability to \textit{Transport Module}.

\subsection{LLM Integration for Semantic Reasoning}
In cases where rule-based scoring cannot definitively determine the suitability of an agent-task pairing, particularly when the task is unclear or novel or a robot is introduced with new capabilities, HVBTA integrates a pre-trained LLM to prompt for a score to evaluate suitability more holistically. The LLM utilizes its semantic understanding to interpret subtle nuances within both the \textit{Task Descriptions} and agent \textit{Capability Profiles}, refining the suitability assessment. An automatically generated prompt focuses the LLM on the specific component(s) whose compatibility is in question, allowing for more flexible assessment of suitability.

\subsection{Voting and Allocation Mechanism}
HVBTA employs a robust voting and allocation mechanism to resolve potential conflicts and balance assignments. This system leverages six distinct voting methods, such as Borda, approval, and majority voting, to aggregate and interpret the scores from the \textit{Suitability Matrix}. This ensures that assignments are aligned with both the agents’ capabilities and the tasks’ requirements, while also considering the overall distribution of tasks across the heterogeneous team. Recognizing that a single, highly capable agent might be the best candidate (highest-scoring) for multiple tasks, HVBTA is capable of delegating tasks efficiently and not relying on the most capable agents. For example, in Table \ref{table:1}, even though \textit{Agent C} scores well on its suitability with both tasks, HVBTA would assign it to \textit{Transport Module} because no other agent scored well on it. \textit{Agent A} would be assigned to \textit{Place Wall Panel} while \textit{Agent B} would remain unassigned.

\subsection{CBS for Path Planning}
After task assignments are finalized, HVBTA integrates CBS for path planning. It plans optimal paths for the agents from their start positions to their designated tasks while avoiding obstacles and collisions with other agents in the physical environment. Path planning finds an optimal sequence of states, locations on a map, to move a robot from one location to another. This integration ensures efficient, collision-free spatial coordination, which is critical for the safe and timely execution of tasks. CBS can dynamically update agents' paths as the environment or an agent's state changes. Although HVBTA does not address task execution, task planning algorithms could also be integrated.

By automating decisions regarding task assignment and coordination that might previously require external intervention, HVBTA addresses key limitations of earlier approaches and enables the creation of efficient and dynamically managed automated workflows on a construction site.

\section{Applying HVBTA to Multi-agent Construction}

The challenges of coordinating heterogeneous MAS for modular construction and site preparation tasks are significant. These tasks require a wide range of capabilities and frequently involve unforeseen requirements that must be addressed through contextual reasoning. HVBTA is particularly well-suited to address these challenges.

\subsection{Handling Heterogeneity} 
HVBTA's emphasis on detailed agent \textit{Capability Profiles} allows the framework to explicitly model the unique strengths, tools, and limitations of each heterogeneous agent in the team. \textit{Task Descriptions} capture the specific requirements of different construction tasks. The \textit{Suitability Matrix} quantitatively compares these profiles and descriptions, enabling the system to identify which agent is best equipped for a given task, whether it requires heavy lifting, fine manipulation, or robust pushing power. For example, a construction site with specialized robots for material transportation, dexterous manipulation, and site monitoring could use HVBTA to coordinate their behavior so that they are assigned to the tasks that each is best suited for, particularly when the robots have diverse skillsets and strengths.

\subsection{Efficient Task Assignment} 
The combination of suitability scoring and multiple voting methods allows HVBTA to efficiently assign tasks across the entire team. This process considers not just the single best agent for a task, but how assignments can be balanced to utilize the entire team effectively, preventing bottlenecks where one agent is assigned multiple tasks while others are idle. The LLM integration provides the flexibility to resolve ambiguous suitability scores, leveraging contextual understanding to make more informed decisions based on subtle factors in agent capabilities or task needs that might not be captured by quantitative scores alone. For example, on a construction site with powerful winds, recently placed modules may need to be anchored to prevent them from becoming unmoored. Although the task lacks explicit physical requirements like in Table \ref{table:1}, the LLM can infer that an agent with strong anchoring capability, weight, and size is going to be best suited for the task by leveraging contextual understanding of construction dynamics.

\subsection{Seamless Coordination and Path Planning} 
The integration of CBS directly addresses the critical need for efficient and collision-free coordination on a busy construction site. Once tasks are assigned, CBS calculates optimal paths for agents to move to their work locations, considering the positions and planned movements of other agents and obstacles. This decentralized path planning component ensures that agents can operate in shared workspaces safely and efficiently, minimizing delays caused by congestion or collisions. CBS's spatio-temporal path planning abilities are vital for automated construction on a busy work site with multiple active agents and obstacles.

\subsection{Adaptability} 
Although construction planning often involves predetermined steps, the dynamic nature of a construction site (e.g., unexpected obstacles, changes in material distribution) requires adaptable agent behavior. HVBTA's integrated approach, particularly with the LLM for nuanced suitability scoring and CBS for dynamic path re-planning, allows the system to delegate tasks efficiently while adapting to changing site conditions, task requirements, or the team of agents.

We are confident that HVBTA will allow multi-agent construction teams to achieve efficient, conflict-free assignment and coordination. This has the potential to facilitate potentially faster and more accurate modular assembly.

\section{Conclusion}

The increasing complexity of MAS and the structured demands of modular construction highlight the need for sophisticated task allocation and coordination frameworks. The Hybrid Voting-Based Task Assignment (HVBTA) framework draws inspiration from human reasoning to efficiently manage heterogeneous automated MAS. HVBTA manages this by rigorously defining agent \textit{Capability Profiles} and \textit{Task Descriptions}, leveraging a \textit{Suitability Matrix}, employing robust voting mechanisms, using an LLM to resolve ambiguities, and incorporating CBS for efficient, collision-free spatio-temporal coordination. This framework is particularly well-suited for the challenges of modular construction and site preparation, enabling efficient task assignment and coordination of agents engaged in tasks like push manipulation and prefabricated assembly. Current work is focused on evaluating HVBTA’s performance across various simulated scenarios involving diverse robotic platforms and task complexities. Ultimately, HVBTA holds significant promise for enhancing the efficiency, safety, and scalability of multi-agent construction.

\bibliographystyle{ieeetr}
\bibliography{bibliography.bib}

\begin{thebibliography}{10}

\bibitem{DAVILADELGADO2019100868}
J.~M. {Davila Delgado}, L.~Oyedele, A.~Ajayi, L.~Akanbi, O.~Akinade, M.~Bilal, and H.~Owolabi, ``Robotics and automated systems in construction: Understanding industry-specific challenges for adoption,'' {\em Journal of Building Engineering}, vol.~26, p.~100868, 2019.

\bibitem{LIM2022104455}
Y.-W. Lim, P.~C. Ling, C.~S. Tan, H.-Y. Chong, and A.~Thurairajah, ``Planning and coordination of modular construction,'' {\em Automation in Construction}, vol.~141, p.~104455, 2022.

\bibitem{xiao2022recent}
B.~Xiao, C.~Chen, and X.~Yin, ``Recent advancements of robotics in construction,'' {\em Automation in Construction}, vol.~144, p.~104591, 2022.

\bibitem{article}
Z.~Chen, J.~Alonso-Mora, X.~Bai, D.~Harabor, and P.~Stuckey, ``Integrated task assignment and path planning for capacitated multi-agent pickup and delivery,'' {\em IEEE Robotics and Automation Letters}, vol.~PP, pp.~1--1, 04 2021.

\bibitem{ali2022considerationstaskallocationhumanrobot}
A.~Ali, D.~M. Tilbury, and L.~P.~R. Jr, ``Considerations for task allocation in human-robot teams,'' 2022.

\bibitem{CHAKRAA2023104492}
H.~Chakraa, F.~Guérin, E.~Leclercq, and D.~Lefebvre, ``Optimization techniques for multi-robot task allocation problems: Review on the state-of-the-art,'' {\em Robotics and Autonomous Systems}, vol.~168, p.~104492, 2023.

\bibitem{taskAllocation1}
L.~Parker, ``Alliance: an architecture for fault tolerant multirobot cooperation,'' {\em IEEE Transactions on Robotics and Automation}, vol.~14, no.~2, pp.~220--240, 1998.

\bibitem{taskAllocation2}
B.~Kalecı, O.~Parlaktuna, M.~Özkan, and G.~Kirlik, ``Market-based task allocation by using assignment problem,'' in {\em 2010 IEEE International Conference on Systems, Man and Cybernetics}, pp.~135--141, 2010.

\bibitem{weiner2025hybridvotingbasedtaskassignment}
D.~Weiner and R.~Korpan, ``Hybrid voting-based task assignment in role-playing games,'' in {\em HRI 2025 Dungeons, Neurons, and Dialogues: Social Interaction Dynamics in Contextual Games Workshop}, 2025.

\bibitem{tang2023largelanguagemodelsincontext}
X.~Tang, Z.~Zheng, J.~Li, F.~Meng, S.-C. Zhu, Y.~Liang, and M.~Zhang, ``Large language models are in-context semantic reasoners rather than symbolic reasoners,'' 2023.

\end{thebibliography}
\end{document}